\pdfoutput=1

\documentclass[11pt]{article}

\usepackage[]{acl}

\usepackage{times}
\usepackage{latexsym}
\usepackage{amsmath}
\usepackage{booktabs}
\usepackage[T1]{fontenc}
\usepackage{graphicx}

\usepackage[utf8]{inputenc}

\usepackage{microtype}

\usepackage{inconsolata}
\usepackage{paralist} 

\title{
Emotion Granularity from Text: \\ 
An Aggregate-Level Indicator of Mental Health  
\vspace{0.115in}}

\author{Krishnapriya Vishnubhotla$^{1,2}$ \and Daniela Teodorescu$^{3}$ \and Mallory J. Feldman$^{4}$ \\ \textbf{Kristen A. Lindquist}$^{4}$ \and \textbf{Saif M. Mohammad}$^{5}$ \\ 
\textsuperscript{1}Department of Computer Science, University of Toronto \\ \textsuperscript{2}Vector Institute, Toronto \\ \textsuperscript{3}Department of Computing Science, University of Alberta\\  \textsuperscript{4}Department of Psychology and Neuroscience, University of North Carolina at Chapel Hill \\ \textsuperscript{5}National Research Council Canada \\}

\begin{document}
\maketitle
\begin{abstract}
We are united in how emotions are central to shaping our experiences; yet, individuals differ greatly in how we each identify, categorize, and 
express emotions.
In psychology, variation in the ability of individuals to differentiate between emotion concepts is called \textit{emotion granularity} (determined through self-reports of one's emotions). High emotion granularity has been linked with 
better mental and physical health;
whereas low emotion granularity has been linked with maladaptive emotion regulation strategies and poor health outcomes. 
In this work, we propose computational measures of emotion granularity derived from temporally-ordered speaker utterances in social media (in lieu of self-reports that suffer from various biases). We then investigate the effectiveness of such text-derived measures of emotion granularity in functioning as markers 
of various mental health conditions (MHCs).
We establish baseline measures of emotion granularity derived from textual utterances, and show that, at an aggregate level,
emotion granularities are significantly lower for 
people self-reporting as having an MHC than for the control population.
This paves the way towards a better understanding of the MHCs, and specifically the role emotions play in our well-being.
\end{abstract}

\section{Introduction}
Emotions play a central role in how we construct meaning and communicate with those around us. 
Yet, individuals vary in their understanding and experience of emotions, or ``emotional expertise'' \cite{Hoemann2021ExpertiseIE}. Some people are able to recognize, identify, and describe what they feel using precise, context-specific terms like guilt, anger, frustration, or helplessness; others tend to use more broad terms to convey a general sense of feeling bad or feeling low. 
\textit{Emotion granularity} (EG), aka \textit{emotion differentiation}, is defined by psychologists as the ability of an individual to experience and categorize emotions in very specific terms \cite{Barrett2001KnowingWY}.
Highly granular individuals have a broad range of highly situated and differentiated emotion concepts, and can reliably describe these concepts using language --- for example, distinguishing between when they are feeling angry vs.\@ when they are feeling sad, or when they are feeling elated from when they are feeling content.

Evidence collected in the last two decades provides consistent support for a link between emotional granularity and mental health \cite{Erbas2014NegativeED, erbas2018don, starr2017feelings, seah2020emotion}, physical health \cite{Hoemann2021ExpertiseIE, bonar2023examining}, and adaptive health behavior \cite{DixonGordon2014APE, kashdan2015unpacking}.
Note that this is different from other findings that study how the prevalence of specific emotions varies with mental health,
(for example, people with depression tend to use more sadness-associated words). 
The link between EG and mental health suggests that there is a fundamental difference in how one \textit{perceives} an emotion (broadly or specifically),
and that in turn can impact their 
mental health.

Typically, 
granularity is measured 
across emotions with the same valence; one can therefore have a measure of \textit{negative emotion granularity}, measured as the 
granularity of negative emotions 
(such as anger, sadness, and fear)
and \textit{positive emotion granularity}, measured 
as the granularity of positive emotions
(such as joy, excitement, and satisfaction). Some works also look at the co-endorsement of emotions that express opposite valence, such as joy and sadness \citep{lindquist2008emotional}.

In psychology and the affective sciences, emotion granularity is often measured using repeated measurements, where individuals are asked to rate the intensity of experiencing certain emotions multiple times over a period of days (e.g., 2--3 times each day for 5 days), i.e, with self-reports of emotions felt.
An individual's emotional granularity is then operationalized as the extent to which multiple emotions are \textit{co-endorsed} over time, i.e, how similarly the emotions are rated across all measurements, using the intraclass correlation coefficient (ICC) \citep{shrout1979intraclass}, which measures the extent to which the emotions co-vary in reports at the aggregate level.
Individuals who tend to frequently rate multiple emotions at the same intensity levels are defined as low in granularity --- the frequent co-endorsement across time indicates that they are failing to differentiate between these
emotions in their reports.
In contrast, individuals high in emotion granularity co-endorse multiple emotions
less frequently over time 
\citep{Tugade2004PsychologicalRA,emo_gran_icc,emo_gran_brain,reitsema2022emotion}.

While prior work in NLP has studied the link between emotions and mental health, these have largely been limited to measuring the prevalence or intensity of positive and negative emotions.
In this work, we, for the first time, propose a way to compute emotion granularity \textit{from the textual utterances of an individual}. 
Our method uses the temporal sequence of the utterances to first construct emotion arcs along multiple emotions, and computes granularity as the correlation of these emotion arcs. We hypothesize that this measure is indicative of the individual consistently expressing the same set of emotions together over a period of time, and can therefore act as a proxy measure of emotional granularity. 

We then study the relationship between aggregate, population-level measures of emotion granularity in text for eight Mental Health Conditions (MHCs), namely attention-deficit hyperactivity disorder (ADHD), anxiety, bipolar disorder, depression, major depressive disorder (MDD), obsessive-compulsive disorder (OCD), postpartum depression (PPD), and post-traumatic stress disorder (PTSD), and compare them to a control group. We use two social media datasets where users have chosen to self-disclose 
their mental health diagnosis \citep{_Singh_Arora_Shrivastava_Singh_Shah_Kumaraguru_2022, eRisk2017, eRisk2018}. We compute emotion granularity metrics for each of these groups to answer the following  questions:\\[-20pt]
\begin{enumerate}
    \item Do measures of emotional granularity differ between the MHCs and the control group?\\[-20pt]
    \item Which measures of emotion granularity are the most effective at differentiating between the MHCs and the control group?\\[-20pt]
    \item Which emotion pairs lead to the greatest difference in granularity between an MHC and the control group?\\[-20pt]
\end{enumerate}
Exploring this line of questions helps us better understand how emotion granularity presents itself in text, whether emotion granularity from text can be a useful tool to study MHCs, and how an MHC impacts our perception of emotions (and perhaps even, how the perception of emotions impacts our mental health).

Our results establish baseline measures of emotion granularity from text, and show that these measures function as reliable indicators, at the aggregate-level, for the presence of many of the mental health conditions we study.
Our work makes an important contribution to the growing wealth of research on textual measures of emotional expression as biosocial markers of MHCs, and has a broader utility in functioning as an additional indicator of the mental well-being of populations.\footnote{The term \textit{biosocial} marker \citep{biosocial} was coined to indicate the crucial role social factors (e.g., socioeconomic status, years of education, bilingualism, etc.) have on quantitative features associated with medical conditions (biomarkers).}
 
All our code will be made available through the project webpage.\footnote{\url{https://github.com/Priya22/emotion-granularity-from-text}}

\section{Related Work}

\subsection{Emotions and Mental Health}
Measures of emotional experience and their patterns of change over time have been extensively studied as markers of mental and physical well-being \cite{lewis2010handbook}. 
The Emotion Dynamics framework in psychology quantifies the patterns with which emotions change over time, allowing researchers to better understand emotional experiences and individual variation \cite{KUPPENS201722}. The framework includes several measures such as the duration, intensity, variability, and granularity of one's emotional experiences. 
Numerous studies in psychology have shown emotion dynamics correlate with overall well-being, mental health, and psychopathology
(the scientific study of mental illness or disorders) \citep{KUPPENS201722,Houben2015,depressedYouth,sperry}.

Emotion granularity in particular is positively associated with adaptive behaviour in adverse conditions --- accurately labeling our emotions can inform us of the right coping strategies to use in different contexts. Individuals with higher emotion granularity tend to use a broader range of strategies to deal with negative emotions, and are more successful at doing so \cite{Barrett2001KnowingWY}. Several studies have shown that emotion granularity is lower in individuals with mental health conditions like bipolar disorder \cite{suvak2011emotional, DixonGordon2014APE}, manic depressive disorder \cite{demiralp2012feeling}, schizophrenia \cite{kring2003broad}, autism spectrum disorder \cite{erbas2013emotion}, and affective disorders like anxiety \cite{seah2020emotion} and depression \cite{starr2017feelings, willroth2020depressive}. Lower granularity is also associated with increased tendencies to engage in maladaptive behaviour, such as alcohol consumption \cite{kashdan2015unpacking, emery2014emotion} and aggression \cite{pond2012emotion}.

Researchers in affective science typically measure emotional granularity through experience sampling methodologies (ESMs), or ecological momentary assessments (EMAs), where individuals (participants) are repeatedly asked to report on their emotional states on several occasions throughout the day, for several days. 
For example, participants may be asked to endorse a series of ten emotion words (e.g., \textit{anger}, \textit{fear}, \textit{happy}, etc.) on a Likert scale across several sampling instances. Emotional granularity would then be computed as the intraclass correlation (ICC) of ratings across sampling instances. A high ICC would suggest that a participant experiences all of the emotions in a similar way across trials (treating them as synonyms for more general affectual states such as ``unpleasantness" or ``pleasantness"), whereas a low ICC would suggest that a participant experienced emotions in a granular and context-specific way.

While emotion granularity is generally measured between emotion categories that are close to each other in the affective space (i.e, express similar valence), the concept of \textit{dialecticism} refers to the co-incidental experience of both negative and positive emotions 
\cite{lindquist2008emotional}. Dialecticism can therefore be operationalized as the co-endorsement of emotion pairs that express positive and negative valence.

\subsection{Language and Mental Health}
Given the limitations of self-report surveys (e.g., limited data coverage and time spans, biases, etc. \citep{Kragel2022}), another approach to measure well-being indicators is through one's language usage. Some well-known linguistic indicators of mental health include the proportion of pronouns used for those with depression \citep{koops}, syntax reduction for anorexia nervosa \citep{Cuteri}, certain lexical and syntactic features for mild cognitive impairment and dementia \citep{CALZA2021101113, Gloria}, and semantic connectedness for schizophrenia \citep{CORCORAN2020158}. 

Recently, another linguistic feature that researchers leveraged for insights into overall well-being, are the emotions expressed in language. Largely, only sentiment has been explored and mainly from social media data (a rich source of language data). 
For example, more negative sentiment was expressed in text by individuals with depression \citep{DeChoudhury,seabrook,De_Choudhury_Gamon_Counts_Horvitz_2021}. Other work has found that suicide watch, anxiety, and self-harm subreddits had markedly lower negative sentiment compared to other mental health subreddits such as Autism and Asperger's \citep{gkotsis-etal-2016-language}. 

\newcite{hipson2021emotion} and \newcite{vishnubhotla-mohammad-2022-tusc} introduced Utterance Emotion Dynamics (UED), a framework to quantify patterns of change of emotional states
associated with \textit{utterances} along a longitudinal (temporal) axis (using data from screenplays and tweets).
\citet{teodorescu-etal-2023-language} found that measures of emotion dynamics from text correlate with various mental health diagnoses. 

These works overall 
show that the average emotion expressed in text and also the characteristics of individual emotion change over time (e.g., variability) are meaningful indicators of well-being. In this work, we explore whether
the degree of co-expression of pairs of emotions in text (emotion granularity) is a meaningful indicator of mental health.

\section{Datasets}
\label{sec:datasets}
We use the Twitter-STMHD dataset \cite{_Singh_Arora_Shrivastava_Singh_Shah_Kumaraguru_2022} for our experiments and also verify our results with a smaller Reddit eRisk \cite{eRisk2017, eRisk2018} dataset. We describe both of them below.

\noindent\textbf{Twitter-STMHD dataset}: 
\citet{_Singh_Arora_Shrivastava_Singh_Shah_Kumaraguru_2022} identified tweeters who self-disclosed as having an MHC diagnosis using carefully constructed regular expression patterns and manual verification. We summarize key details on the dataset creation process in Appendix \ref{app:twitter_dataset}. 
The control group consists of users identified from a random sample of tweets (who posted during approximately the same time period as the MHC tweets). These tweeters did not post any tweets meeting the MHC regex described above.
Additionally, users who had any posts about mental health discourse were removed from the control group. 
Note that this process does not guarantee that users in the control group did not have an MHC diagnosis, but rather the group as a whole may have very few tweeters from these MHC groups.
The number of users in the control group was selected to match the size of the depression dataset, which had the largest number of users.

For the final set of users, four years of tweets were collected for each user: two years before self-reporting a mental health diagnosis and two years after. For the control group, tweets were randomly sampled from between January 2017 and May 2021 (same date range as the other MHC classes).

\noindent \textbf{Reddit eRisk dataset}\label{sec:eRisk_dataset}:   
To further add to our findings, we also included the eRisk 2018 dataset \cite{eRisk2017, eRisk2018} in our experiments. 
It consists of users who self-disclosed as having depression on Reddit (expressions were manually checked), and a control group (individuals were randomly sampled). 
The dataset includes several hundred posts per user, over approximately a 500-day period. We combined users and their instances from both the training set (which is from the eRisk 2017 task \cite{eRisk2017}) and the test set (from the eRisk 2018 task \cite{eRisk2018}). 

\subsection{Preprocessing}
\label{sec:preprocessing}

We further preprocessed both the Twitter-STMHD dataset and the eRisk dataset for our experiments (Section \ref{sec:experiments}), as
we are specifically interested in the relationship between emotion granularity and each disorder.
Several users self-reported as being diagnosed with more than one disorder, referred to as \textit{comorbidity}. 
We found a high comorbidity rate between users who self-reported as having anxiety and depression, as is also supported in the literature \citep{pollack2005comorbid,gorman,PMID:15014592,Cummings_2014}.
Since we wanted to focus on each MHC separately (and not on co-morbidity) we only considered users who self-reported as having one MHC. 
We also performed the following preprocessing steps: 

\begin{compactitem}
\item  We only considered posts in English.
\item We filtered out posts that contained URLs (the text in such posts is often not self-contained).
\item We removed retweets (identified through tweets containing `RT', `rt'). This is to focus exclusively on texts written by the user.
\item 
To ensure that we did not include users that post very infrequently or very frequently, we excluded 
users
based on the number of posts per individual. 
We discarded data from those who either had less than 100 posts (as was similarly done in \citet{vishnubhotla-mohammad-2022-tusc}) 
and
those who had posted more than 1.5 times the interquartile range above quartile three (75th percentile) 
of the control group.\footnote{The interquartile range is from the 25th to 75th percentile.}
\end{compactitem}
 \noindent Table \ref{tab:twitter-stmhd_dataset_descriptives} shows key details of the filtered Twitter-STMHD and Reddit eRisk datasets.

\begin{table}[t]
\centering
{\small
\begin{tabular}
{lrrr}
\hline
\textbf{Dataset, Group} &
\textbf{\#people} & 
\textbf{\raggedright Av.\@ \#posts} & 
\multicolumn{1}{p{1.5cm}}{\textbf{\raggedright Av.\@ \#tokens \\ per post}} \\ 
\hline
\textit{Twitter} &  &  &  \\
$\;\;\;$ MHC & 19,324  & 2,590.48 & 17.59 \\
$\;\;\;$  $\;\;\;$ ADHD & 6,356  & 2,497.43 & 17.46 \\
$\;\;\;$  $\;\;\;$ Anxiety & 3,036 &  2,921.05 & 17.46 \\
$\;\;\;$  $\;\;\;$ Bipolar & 1,061  &  2,820.17 & 17.32 \\
$\;\;\;$  $\;\;\;$ Depression & 4,855 &  2,526.62 & 16.75 \\
$\;\;\;$  $\;\;\;$ MDD & 219 & 2,640.69 & 16.40 \\ 
$\;\;\;$  $\;\;\;$ OCD & 1,009 &  2,388.73 & 18.38 \\ 
$\;\;\;$  $\;\;\;$ PPD & 179 & 2,581.19 & 19.18 \\ 
$\;\;\;$  $\;\;\;$ PTSD & 2,609 & 2,533.85 & 19.41 \\
$\;\;\;$  Control & 6,001 & 2,420.50 & 16.16 \\
\textit{Reddit} &  &  &\\
$\;\;\;$  Depression & 112 & 556.57 & 47.22 \\
$\;\;\;$  Control & 907 & 665.00 & 41.09 \\
\hline
\end{tabular}
}
 \vspace*{-1mm}
\caption{ The number of users in each MHC, 
the average number of posts per user, and the average number of tokens per post in the preprocessed version of the Twitter-STMHD and Reddit eRisk datasets.}
 \vspace*{-2mm}
\label{tab:twitter-stmhd_dataset_descriptives}
\end{table}

\section{Emotion Granularity from Text}
\label{sec:experiments}
The core metric that we want to capture from the text utterances of an individual is emotion granularity---what psychologists term the ``co-endorsement" of pairs of emotions. Analogous to their operationalization of granularity in terms of the Intra-Class Correlation (ICC) of repeated emotion intensity measurements along emotion adjectives, we use textual utterances to derive a temporal sequence of emotion states, referred to as an \textit{emotion arc} for the speaker (section 4.1), and operationalize granularity as the correlations of these arcs. We construct emotion arcs for multiple emotions, for each user in the MHC groups and the control group. 

\noindent\textbf{Emotion Dimensions: }
A key requirement of our computational method is that we must be able to quantify the emotional score of a text along a selected emotion dimension. We are therefore limited by the resources and models available to compute such a score for an emotion dimension. 

Here, keeping in mind the necessity of including multiple emotion dimensions that are similarly-valenced, we work with the eight emotions represented in the NRC Emotion Intensity Lexicon: anger, anticipation, disgust, fear, joy, sadness, surprise, and trust \cite{LREC18-AIL}. 

We partition these emotions into three groups based on the valence association: joy and trust are in the \textbf{positive valence} group; anger, sadness, fear, and disgust are in the \textbf{negative valence} group; and anticipation and surprise are in the \textbf{variable valence} group. The distinction for surprise and anticipation is necessary because specific instances of these emotions can have a positive or a negative connotation (e.g., a good or a bad surprise).

\subsection{Constructing Emotion Arcs}
\label{sec-arcs}
We order the utterances for each user based on timestamp information in the metadata.
We construct emotion arcs for the temporal sequence of utterances of each user, along each of the eight emotions, in two ways pertaining to different window sizes. This is to make sure that the results are largely robust even when varying the window size to some extent.\\[-22pt]
\paragraph{Utterance-level Window:}  Emotion scores (for each emotion category) are computed for each utterance (i.e, tweet or Reddit post). Here, an utterance is  assumed to represent the speaker's emotion state at a particular point in time (analogous to sampling instances).
The sequence of utterance emotion scores for a user 
forms their temporal emotion arc.\footnote{The frequency and time of posting often differs between users, but we ignore that for now.}\\[-22pt]
\paragraph{Word-Count based Window:} Here, the emotion score at a point in time is computed for a window of words (say, 100 words) that are uttered around that point, and the window is moved forward by a fixed step size (say, 1 word at a time) to obtain the emotion score for the next time step. In prior work on constructing emotion arcs from temporally-ordered text, such sliding windows are usually employed to ensure smoother arcs that more accurately capture the flow of emotions over time.

\citet{teodorescu2023evaluating} conducted extensive quantitative evaluations of several hyperparameters involved in emotion arc construction, on datasets from diverse domains (including tweets) annotated with emotion scores. We follow many of their recommendations to construct emotion arcs for the utterances of each of our users.

The texts are tokenized using the \texttt{twokenizer}\footnote{\url{https://github.com/myleott/ark-twokenize-py}} package to obtain a similarly-ordered sequence of words.
Emotion scores are computed with window sizes of 100 words and 500 words each, and the window is moved forward by one word at each timepoint to obtain a series of overlapping emotion scores. 

\noindent\textbf{Emotion scoring method}:
Keeping in mind the necessity of an \textit{interpretable} method of emotion scoring, we use word--emotion lexicons to compute the emotion scores of text spans. For each window, the emotion scores of its constituent words are averaged to obtain the window-level score for that emotion. \citet{teodorescu2023evaluating} showed that emotion arcs constructed with lexicon-based scoring methods, when used with sliding window sizes of 100 instances or more, can mimic the ground-truth emotion arcs with an accuracy of 0.9 or more.

Word--emotion scores are obtained from the NRC Emotion Intensity Lexicon, which associates close to 10,000 English words with a real-valued score between 0 and 1 for each dimension. A score of 0 indicates that the word has little to no association with that particular emotion, and a score of 1 indicates a high association.

\noindent\textbf{Qualitative Checks on Emotion Lexicons}:
Lexicon-based methods for constructing emotion arcs are reliable and interpretable; however, it is good practice to 
modify the lexicon to the specific domain of use, in order to account for 
terms that are expected to be used in the target domain in a sense different from the predominant word sense \cite{mohammad-2023-best}. 

We identify and remove words and bigrams whose usage on Twitter (and sometimes more colloquially) is markedly different from the predominant word sense annotated in the lexicons, such as \textit{like} and \textit{chaotic evil}. We also remove words and bigrams that are explicitly associated with mental health, such as \textit{anxious, disorder} and \textit{panic attack}. Though our EG metric does not explicitly rely on the presence of such terms to find associations with MHCs, we remove them in order to capture more fundamental differences in emotional expression between users in the MHC groups and the control group. The full list of stopwords is in Appendix \ref{app:lex_words_removed}.

\noindent\textbf{Hyperparameters:} We additionally make the following choices of hyperparameters for constructing and comparing a pair of emotion arcs:
\begin{itemize}
    \item For a given pair of emotions, we drop all emotion terms that are common to the two lexicons before constructing their emotion arcs. This ensures that we are not using words associated with both emotions, giving us a clearer indication of co-endorsement.
    \item An utterance (or window) with no emotion terms from a particular emotion lexicon is assigned a score of 0. An alternative is to assign them a score of \texttt{nan}, in which case they are not considered a part of the emotion arc.\footnote{We do not observe any major changes to our results based on these hyperparameter choices.} 
\end{itemize} 
A visualization of the emotion arcs obtained using the utterance-level window for a sampled user from the Twitter-STMHD dataset is presented in Appendix \ref{app:viz}.

\subsection{Quantifying Emotion Granularity}
\label{sec:quant_emo_gran}
We 
compute 
the emotion granularity metric as the \textit{negative} of the Spearman correlation between each pair of emotions arcs, for each user.\footnote{We choose Spearman correlation as it is rank-based as compared to Pearson correlation which utilizes the raw-values.} A high correlation between two arcs indicates that the speaker is consistently and repeatedly expressing the two emotions concurrently; we hypothesize that this is an indicator of a lower ability to \textit{differentiate} between the two emotions, and therefore a lower emotion granularity.\footnote{We choose to use Spearman correlation over ICC-based metrics because the emotion scores that we extract from textual utterances are a \textit{relative} indicator of the intensity of the emotion, and not an absolute measure. Further, these scores cannot be directly compared \textit{across} different emotions in terms of absolute intensity (a score of 0.9 for anger may not equate to the same level of anger as a score of 0.9 would for joy) due to differences in how overtly different emotions are expressed via language.}

For each person, we average the correlation scores between emotion pairs 
in the different valence groups to obtain the following measures of emotion granularity (EG):
\begin{compactitem}
    \item $EG(pos)$: The negative of (i.e., $-1$ times) the average of the correlation scores between each of the pairs of emotions in the \textbf{positive valence} group (joy--trust).
    \item $EG(neg)$: The negative of the average of the correlation scores between each of the pairs of emotions in the \textbf{negative valence} group (anger--fear, fear--disgust, etc.).
    \item $EG(var)$: The negative of the average of the correlation scores between each of the pairs of emotions in the \textbf{variable valence} group (surprise--anticipation).
    \item $EG(overall)$: Overall emotion granularity, measured as the negative of the average of the correlation scores between emotion pairs whose constituents are in the \textbf{same group}. Here, the average is taken across all of the emotion pairs drawn from the positive valence group, the negative valence group, and the variable valence group.
    \item $EG(cross)$: Emotion granularity of cross-group emotion pairs. That is, the negative of the average of the correlation scores between emotion pairs whose constituents come from {\bf different groups}. This measure to some extent quantifies the amount of \textit{dialecticism} (expressing both positive and negative emotions in a narrow window of time); however, note that EG(cross) also includes emotions that express variable valence (surprise and anticipation), rather than only considering positive--negative valence emotion pairs.
    
\end{compactitem}

We consider $EG(overall)$ to be the 
bottom line measure of emotion granularity for a user (analogous to that used in psychology studies). Note that cross-group pairs are not included in this measure.

\begin{table*}[t]
\centering
{\small
  \begin{tabular}{lcccccccc}
\hline
\textbf{Dataset, MHC--Control} &
 $IC(n)$ & $IC(v)$ &
 $EG(pos)$ & 
  $EG(neg)$ &
  $EG(var)$ &
  $EG(cross)$ &
  $EG(overall)$ \\
 \hline
\textit{Twitter-STMHD} \\
$\;\;\;$ ADHD--control & -- & -- & lower & lower & lower & lower & lower \\  
$\;\;\;$ Anxiety--control & -- & -- & lower & lower & lower & lower & lower \\   
$\;\;\;$ Bipolar--control & -- & -- & lower & lower & lower & lower & lower\\
$\;\;\;$ MDD--control & -- & -- & lower & -- & -- & lower & lower\\     
$\;\;\;$ OCD--control & -- & -- & lower & lower & lower & lower & lower\\     
$\;\;\;$ PPD--control & -- & -- & -- & lower & -- & -- & --\\     
$\;\;\;$ PTSD--control & -- & -- & lower & lower & lower & lower &  lower\\
$\;\;\;$ Depression--control & -- & -- & lower & lower & lower & lower & lower \\
\textit{Reddit eRisk} \\
$\;\;\;$ Depression--control & -- & -- & lower & lower & -- & lower & lower \\ 
\hline 
\end{tabular}
}
\caption{
The difference in emotion granularity between each MHC group and the control. A significant difference is indicated by the word `lower' or `higher', indicating the direction of the difference in granularity. 
}
 \vspace*{-3mm}
\label{tab:arrows_all_emotions_no_comorbid_zero-50-25_ex}
\end{table*}

\section{Emotion Granularity and Mental Health}
\label{sec:analyses}
We now test if there are significant differences between the emotion granularities of each of the MHC groups and the control group, using t-tests. We first limit the users in each group by placing thresholds on (a) the number of user tweets with a valid emotion score (set to a minimum of 50), and (b) the number of unique lexicon terms used in their tweets (set to a minimum of 25). These thresholds ensure that we are drawing inferences based on users with \textit{valid} emotion arcs, with sufficient lexicon coverage and temporal information. 

We performed independent t-tests to compare emotion granularities between each of the MHCs and the control group, for each emotion group, using the \texttt{SciPy} library \cite{2020SciPy-NMeth}. To correct for multiple comparisons (eight tests performed for each MHC per emotion granularity group), we used the Benjamini–Hochberg procedure in the \texttt{statsmodels} library \cite{seabold2010statsmodels}. Further details on the data assumptions for t-tests are in Appendix \ref{app:statistical_assumptions}.

\subsection{Term Specificity as a Control}
\label{sec:word_specificity}
Lower emotion granularity occurs when, for a person, the concepts of the relevant emotions are so broad (and non-specific) that their meanings overlap substantially.
This work is testing the hypothesis of whether people who have self-disclosed as having an MHC have lower emotion granularity than those that do not. 
However, another plausible hypothesis is that 
people in a particular group (e.g., MHC or the control) tend to 
use more specific words overall. Doing so would imply a higher specificity (i.e, a higher granularity) in their usage of \textit{all words}, and that the high granularity of emotion words is simply a by-product of their general style of speaking (or posting online). 

To ensure that the level of word specificity does not differ between MHCs and the control group and act as a confounder for our measure of emotion granularity,
we performed a control experiment. We compute the average \textit{information content} of the noun and verb terms in the posts of users in each group, and use this as a measure of the specificity of their language. 
We use the metric proposed in \citet{Resnik1995UsingIC}, and implemented in the NLTK WordNet library,\footnote{\url{https://github.com/nltk/wordnet}} which combines information about the depth of the term in the WordNet tree hierarchy and its frequency of occurrence in a large corpus (here, the Brown corpus) to compute an information content score \cite{10.1145/219717.219748}. 
We then compute the following measures of term specificity for each user:
\begin{compactitem}
    \item $IC(n)$: The information content score for all nouns 
    is averaged across all posts of each individual in each group.\\[-10pt]
    \item $IC(v)$: The information content score for all verbs 
    is averaged across all posts of each individual in each group.
\end{compactitem}
Statistical tests for significant differences are similarly performed as described above (Section \ref{sec:analyses}).

\section{Results}
\label{sec:results}
In Table \ref{tab:arrows_all_emotions_no_comorbid_zero-50-25_ex} we report the statistical results from the pairwise comparisons between each MHC and the control group, for the control experiment on general term specificity as well as emotion granularity,  when scores are computed at the utterance-level. 

All statistically significant differences between an MHC and the control group are described as either `higher' or `lower', and a dash (--) for no statistical difference. 
A `lower' value in a cell indicates that the MHC (rows) has lower emotion granularity (or lower term specificity) than the control group, i.e., higher correlation between emotion pairs in that group (columns); `higher' indicates the MHC has higher emotion granularity (or higher term specificity) than the control group (i.e., lower correlation between emotion pairs in that group). In Table \ref{tab:group_emo_corr} in the Appendix, we also report the absolute Spearman correlation scores for each group.
Below we summarize the results for each column.

\subsection{Emotion Granularity as an Indicator of MHCs}

\noindent\textbf{IC(n) and IC(v):} We do not see any significant differences in the information content of noun and verb terms ($IC(n)$ and $IC(v)$) between MHCs and the control group. This indicates that no group tends to use more specific or less specific language in general when posting on these platforms. Details on the statistical results are shown in Appendix \ref{app:word_specificity}.

\noindent\textbf{EG(pos):} All MHCs except for PPD had significantly lower positive emotion granularity than the control group  (which had similar granularity compared to the control group). 
That is, tweeters in these MHC groups (ADHD, Anxiety, etc.) consistently expressed multiple positive emotions concurrently, more so than the control group. 

\noindent\textbf{EG(neg):} All MHCs except MDD had significantly lower negative emotion granularity than the control group, in both datasets. 
Thus, tweeters in these MHCs were generally not 
differentiating between the negative emotions of anger, disgust, fear, and sadness, as well as the control group.

\noindent\textbf{EG(var):} Tweeters in the ADHD, Anxiety, Bipolar, OCD, PTSD, and Depression (Twitter-STMHD) had significantly lower variable emotion granularity than the control group (i.e., these groups generally differentiated between surprise and anticipation less than the control group). 

\noindent\textbf{EG(overall):} All MHCs
except PPD 
had significantly lower emotion granularity for emotion categories that express the same valence (the mixed valence emotions of surprise and anticipation are also included here). Tweeters in these groups are therefore expressing multiple close emotions frequently with one another -- more so than the control group. 

\noindent\textbf{EG(cross):} 
All MHCs except PPD
had significantly lower granularity between emotion pairs that come from different valence groups. This indicates that positive and negative emotions are expressed together more frequently by tweeters in these groups compared to the control, as well as emotions like joy (positive valence) and surprise (variable valence).

\noindent \textbf{\textit{Discussion}:} These results demonstrate that our measures of emotion granularity from text are consistently lower for users in the MHC groups compared to the control. The term specificity results also tell us that it is the specificity of \textit{emotion word usage} in particular that is differentiating MHCs from the control group.

Aligning with self-report studies in psychology, the emotion granularity between negative-valence emotions is lower for most (7 out of 8) MHCs in our datasets with utterance-level operationalizations. Positive emotion granularity is also correlated with many of the MHCs (7 out of 8 disorders). 
In general, the granularity of emotional expression between within-group emotion pairs is lower for all MHC groups compared to the control in both datasets, except PPD. 
This is in line with both the theoretical and conceptual links established in the psychology literature on emotion granularity and mental health: the ability to better differentiate between emotion concepts that are close to one another leads to more adaptive health behaviour.

While emotion pairs from differently-valenced emotion groups are not usually operationalized in affective science experiments, we find that this measure is also significantly lower in many MHCs. Further investigations into what the concurrent expression of positive and negative emotions means, for emotion granularity and emotion dynamics in general, are interesting research directions.

\noindent\textbf{Variation with hyperparameter choices}: We observe only minor variations from the results reported in Table \ref{tab:arrows_all_emotions_no_comorbid_zero-50-25_ex} when the hyperparameters described in Section \ref{sec-arcs} for emotion arc construction were changed -- less than 10\% of the cells differed in their values across all variations. We provide a more detailed report in Appendix \ref{app:hyper1}.

\subsection{Additional Window Sizes} 
We also examined how the measures of differences in emotion granularity between MHCs and the control change when we compute emotion scores with larger window sizes.

Many of the utterance-level outcomes are replicated for negative, positive, and overall emotion granularity with window sizes 100 and 500. Some measures are no longer significantly different between certain MHCs and the control. We also find that $EG(cross)$ is \textit{higher} for certain MHCs (Anxiety, PPD, PTSD, Depression in Twitter-STMHD) when compared to the control, i.e, users in the control group are expressing negative and positive emotions together more frequently than those in the MHCs.

With larger window sizes, we end up capturing emotions expressed by the individual over longer time spans (tweets posted over the span of several hours or days), rather than co-endorsement at the same time. We hypothesize that these effects of dialecticism, where the control group has a higher co-occurrence of cross-valence group emotions, are capturing the extent to which users balance negative emotions with positive emotions (and vice versa).
The consistent effects with 100 and 500-word windows, and for several MHCs, makes this a promising area for future work.
All emotion granularity measures with window sized 100 and 500 are reported in Appendix \ref{app:emo_granularity_diff_window_sizes}.

\subsection{Individual Emotion Pairs}
In order to understand \textit{which} emotion pairs are expressed together more frequently (resulting in lower emotion granularity), we performed the same significance tests as before between MHCs and the control for correlation scores between all individual emotion pairs. We found that:
\begin{compactitem}
 
    \item Seven out of the eight MHCs in the Twitter-STMHD dataset had a lower granularity (a higher correlation) for anger--disgust (except PPD) and anger--sadness (except MDD) in the negative valence group.
    \item All eight MHCs had a lower emotion granularity (higher correlation) between multiple cross-group emotion pairs, notably those involving the mixed-valence emotions of anticipation and trust.
    \item Contrary to trends, the Bipolar group had a \textit{higher} emotion granularity (i.e, a lower correlation of emotion arcs) for the cross-group emotion pairs of anger--joy and fear--joy.
\end{compactitem}
\noindent Detailed results for each of the emotion pairs and all MHCs are in Appendix \ref{app:emo_gran_emo_pairs}, Table \ref{tab:arrows_all_no_comorbid_v2}. 

\noindent \textbf{Discussion:} 
While lower granularity among certain emotion pairs consistently function as indicators of all MHCs, we also see a few instances where MHCs (specifically Bipolar disorder) have a higher granularity between the emotions when compared to the control. These findings are of interest to researchers studying the links between how emotions are expressed in text, and how they vary with different MHCs. 

\section{Conclusion}
In this work, we operationalized for the first time a computational measure of emotion granularity that can be derived from the textual utterances of individuals. We applied this measure to two social media datasets of posts from individuals who have self-disclosed as having an MHC. Our findings showed that our measure of negative emotion granularity is significantly lower for 7 out of the 8 MHC groups under consideration when compared to a control group, at an aggregate-level. Also, all MHCs 
except for PPD 
had lower overall emotion granularity (and lower positive emotion granularity) compared to the control group. Our work makes an important contribution towards deriving aggregate-level indicators of emotional health from the large amounts of utterance data available on social media platforms. We hope this opens up an avenue of future work to explore emotional expression in text and mental health.

\section*{Limitations}
Our work uses the social media utterances of individuals to derive measures of emotional expression that, at an aggregate level, are found to correlate with multiple mental health conditions. While we use datasets that were compiled by other researchers in the field, we stress that they may not be representative of the general population.
Our methods therefore cannot be directly applied to make inferences on other datasets without a careful experimental validation first. The datasets we study rely on self-disclosures made on social media platforms; it is possible that users report only one such MHC but are diagnosed with others, or that they misrepresent their diagnoses.
Further, the users in the control groups may include those who have chosen to simply not self-disclose on these platforms. 
This can occur due to many reasons, like social desirability \cite{latkin2017relationship} or impression management \cite{tedeschi2013impression}. 
Nonetheless, since we draw inferences at an aggregate level, the methods used can overcome some amount of noisy data.

The set of emotions that we have considered in our measurement of emotion granularity are also limited to those for which we can computationally obtain text-derived emotion scores. These eight emotions do not represent the wide range of emotion concepts that exist and are experienced and expressed by us with language, and future research can attempt to expand our operationalization to more emotion concepts. 
It should be noted though, that past psychology studies on emotion granularity have also tended to explore small sets of emotions, largely because it is cumbersome to ask users about how they feel for a large set of emotions.

The emotion lexicons that we use are some of the largest that exist with wide coverage and large number of annotators (thousands of people as opposed to just a handful). 
However, no lexicon can cover the full range of linguistic and cultural diversity in emotion expression. 
The lexicons are largely restricted to words that are most commonly used in Standard American English and they capture emotion associations as judged by American native speakers of English. See \citet{mohammad-2023-best} for a discussion of the limitations and best-practises in the use of emotion lexicons.

Lastly, further work should explore if the relationships we found hold around various social factors such as age, region, language, etc. As we focus on English text, and the region of users is not known (some information could be extracted from user profiles in the Twitter-STMHD dataset however it is fairly noisy), conclusions should be drawn cautiously across various sociolinguistic factors.\\\\
\section*{Ethics Statement}
Our approach, as with all data-driven models of determining indicators of mental health, should be considered as \textit{aggregate-level indicators}, rather than biomarkers for individuals  \cite{Guntuku2017DetectingDA}. We do not attempt to predict the presence of MHCs for individual users at any stage of the process. These measures should also not be taken as standalone indicators of mental health or mental wellness, even at the population level, but rather as an additional metric that can be used in conjunction with other population-level markers, and with the expertise of clinicians, psychologists, and public health experts.

Individuals vary considerably in how, and how well, they express their internal emotional states using language. Our method of assessing the emotional states of users based on their utterances may miss several linguistic cues of emotion expression, and may not account for individual variation or the extent to which these emotions are expressed on social media.
The emotionality of one's language may also be conveying information about the emotions of the speaker, the listener, or something or someone else mentioned in the utterances. See further discussions of ethical considerations when using computational methods for affective science in \citet{mohammad-2023-best, mohammad-2022-ethics-sheet}. 


\bibliography{custom}

\begin{thebibliography}{59}
\expandafter\ifx\csname natexlab\endcsname\relax\def\natexlab#1{#1}\fi

\bibitem[{Barrett et~al.(2001)Barrett, Gross, Christensen, and
  Benvenuto}]{Barrett2001KnowingWY}
Lisa~Feldman Barrett, James~Jonathan Gross, Tamlin~Conner Christensen, and
  Michael Benvenuto. 2001.
\newblock \href {https://api.semanticscholar.org/CorpusID:6487582} {Knowing
  what you're feeling and knowing what to do about it: Mapping the relation
  between emotion differentiation and emotion regulation}.
\newblock \emph{Cognition and Emotion}, 15:713 -- 724.

\bibitem[{Bonar et~al.(2023)Bonar, MacCormack, Feldman, and
  Lindquist}]{bonar2023examining}
Adrienne~S Bonar, Jennifer~K MacCormack, Mallory~J Feldman, and Kristen~A
  Lindquist. 2023.
\newblock Examining the role of emotion differentiation on emotion and
  cardiovascular physiological activity during acute stress.
\newblock \emph{Affective Science}, pages 1--15.

\bibitem[{Calzà et~al.(2021)Calzà, Gagliardi, {Rossini Favretti}, and
  Tamburini}]{CALZA2021101113}
Laura Calzà, Gloria Gagliardi, Rema {Rossini Favretti}, and Fabio Tamburini.
  2021.
\newblock \href {https://doi.org/https://doi.org/10.1016/j.csl.2020.101113}
  {Linguistic features and automatic classifiers for identifying mild cognitive
  impairment and dementia}.
\newblock \emph{Computer Speech \& Language}, 65:101113.

\bibitem[{Corcoran et~al.(2020)Corcoran, Mittal, Bearden, {E. Gur}, Hitczenko,
  Bilgrami, Savic, Cecchi, and Wolff}]{CORCORAN2020158}
Cheryl~M. Corcoran, Vijay~A. Mittal, Carrie~E. Bearden, Raquel {E. Gur}, Kasia
  Hitczenko, Zarina Bilgrami, Aleksandar Savic, Guillermo~A. Cecchi, and
  Phillip Wolff. 2020.
\newblock \href {https://doi.org/https://doi.org/10.1016/j.schres.2020.04.032}
  {Language as a biomarker for psychosis: A natural language processing
  approach}.
\newblock \emph{Schizophrenia Research}, 226:158--166.
\newblock Biomarkers in the Attenuated Psychosis Syndrome.

\bibitem[{Cummings et~al.(2014)Cummings, Caporino, and Kendall}]{Cummings_2014}
Colleen~M. Cummings, Nicole~E. Caporino, and Philip~C. Kendall. 2014.
\newblock \href {https://doi.org/10.1037/a0034733} {Comorbidity of anxiety and
  depression in children and adolescents: 20 years after.}
\newblock \emph{Psychological Bulletin}, 140(3):816--845.

\bibitem[{Cuteri et~al.(2022)Cuteri, Minori, Gagliardi, Tamburini, Malaspina,
  Gualandi, Rossi, Moscano, Francia, and Parmeggiani}]{Cuteri}
Vittoria Cuteri, Giulia Minori, Gloria Gagliardi, Fabio Tamburini, Elisabetta
  Malaspina, Paola Gualandi, Francesca Rossi, Milena Moscano, Valentina
  Francia, and Antonia Parmeggiani. 2022.
\newblock \href {https://doi.org/10.1007/s40519-021-01273-7} {Linguistic
  feature of anorexia nervosa: a prospective case-control pilot study}.
\newblock \emph{Eating and weight disorders : EWD}, 27(4):1367—1375.

\bibitem[{De~Choudhury et~al.(2013)De~Choudhury, Counts, and
  Horvitz}]{DeChoudhury}
Munmun De~Choudhury, Scott Counts, and Eric Horvitz. 2013.
\newblock \href {https://doi.org/10.1145/2464464.2464480} {Social media as a
  measurement tool of depression in populations}.
\newblock In \emph{Proceedings of the 5th Annual ACM Web Science Conference},
  WebSci '13, page 47–56, New York, NY, USA. Association for Computing
  Machinery.

\bibitem[{De~Choudhury et~al.(2021)De~Choudhury, Gamon, Counts, and
  Horvitz}]{De_Choudhury_Gamon_Counts_Horvitz_2021}
Munmun De~Choudhury, Michael Gamon, Scott Counts, and Eric Horvitz. 2021.
\newblock \href {https://doi.org/10.1609/icwsm.v7i1.14432} {Predicting
  depression via social media}.
\newblock \emph{Proceedings of the International AAAI Conference on Web and
  Social Media}, 7(1):128--137.

\bibitem[{Demiralp et~al.(2012)Demiralp, Thompson, Mata, Jaeggi, Buschkuehl,
  Barrett, Ellsworth, Demiralp, Hernandez-Garcia, Deldin
  et~al.}]{demiralp2012feeling}
Emre Demiralp, Renee~J Thompson, Jutta Mata, Susanne~M Jaeggi, Martin
  Buschkuehl, Lisa~Feldman Barrett, Phoebe~C Ellsworth, Metin Demiralp, Luis
  Hernandez-Garcia, Patricia~J Deldin, et~al. 2012.
\newblock Feeling blue or turquoise? emotional differentiation in major
  depressive disorder.
\newblock \emph{Psychological science}, 23(11):1410--1416.

\bibitem[{Dixon-Gordon et~al.(2014)Dixon-Gordon, Chapman, Weiss, and
  Rosenthal}]{DixonGordon2014APE}
Katherine~L. Dixon-Gordon, Alexander~L. Chapman, Nicole~H. Weiss, and
  M.~Zachary Rosenthal. 2014.
\newblock \href {https://api.semanticscholar.org/CorpusID:26379855} {A
  preliminary examination of the role of emotion differentiation in the
  relationship between borderline personality and urges for maladaptive
  behaviors}.
\newblock \emph{Journal of Psychopathology and Behavioral Assessment},
  36:616--625.

\bibitem[{Emery et~al.(2014)Emery, Simons, Clarke, and
  Gaher}]{emery2014emotion}
Noah~N Emery, Jeffrey~S Simons, C~Joseph Clarke, and Raluca~M Gaher. 2014.
\newblock Emotion differentiation and alcohol-related problems: The mediating
  role of urgency.
\newblock \emph{Addictive Behaviors}, 39(10):1459--1463.

\bibitem[{Erbas et~al.(2013)Erbas, Ceulemans, Boonen, Noens, and
  Kuppens}]{erbas2013emotion}
Yasemin Erbas, Eva Ceulemans, Johanna Boonen, Ilse Noens, and Peter Kuppens.
  2013.
\newblock Emotion differentiation in autism spectrum disorder.
\newblock \emph{Research in Autism Spectrum Disorders}, 7(10):1221--1227.

\bibitem[{Erbas et~al.(2018)Erbas, Ceulemans, Kalokerinos, Houben, Koval, Pe,
  and Kuppens}]{erbas2018don}
Yasemin Erbas, Eva Ceulemans, Elise~K Kalokerinos, Marlies Houben, Peter Koval,
  Madeline~L Pe, and Peter Kuppens. 2018.
\newblock Why i don’t always know what i’m feeling: The role of stress in
  within-person fluctuations in emotion differentiation.
\newblock \emph{Journal of personality and Social Psychology}, 115(2):179.

\bibitem[{Erbas et~al.(2014)Erbas, Ceulemans, Pe, Koval, and
  Kuppens}]{Erbas2014NegativeED}
Yasemin Erbas, Eva Ceulemans, Madeline~Lee Pe, Peter Koval, and Peter Kuppens.
  2014.
\newblock \href {https://api.semanticscholar.org/CorpusID:24782515} {Negative
  emotion differentiation: Its personality and well-being correlates and a
  comparison of different assessment methods}.
\newblock \emph{Cognition and Emotion}, 28:1196 -- 1213.

\bibitem[{Gagliardi and Tamburini(2021)}]{Gloria}
Gloria Gagliardi and Fabio Tamburini. 2021.
\newblock \href {https://doi.org/10.1418/101111} {Linguistic biomarkers for the
  detection of mild cognitive impairment}.
\newblock \emph{Lingue e linguaggio, Rivista semestrale}, (1/2021):3--31.

\bibitem[{Gkotsis et~al.(2016)Gkotsis, Oellrich, Hubbard, Dobson, Liakata,
  Velupillai, and Dutta}]{gkotsis-etal-2016-language}
George Gkotsis, Anika Oellrich, Tim Hubbard, Richard Dobson, Maria Liakata,
  Sumithra Velupillai, and Rina Dutta. 2016.
\newblock \href {https://doi.org/10.18653/v1/W16-0307} {The language of mental
  health problems in social media}.
\newblock In \emph{Proceedings of the Third Workshop on Computational
  Linguistics and Clinical Psychology}, pages 63--73, San Diego, CA, USA.
  Association for Computational Linguistics.

\bibitem[{Gorman(1996)}]{gorman}
Jack~M. Gorman. 1996.
\newblock \href
  {https://doi.org/https://doi.org/10.1002/(SICI)1520-6394(1996)4:4<160::AID-DA2>3.0.CO;2-J}
  {Comorbid depression and anxiety spectrum disorders}.
\newblock \emph{Depression and Anxiety}, 4(4):160--168.

\bibitem[{Guntuku et~al.(2017)Guntuku, Yaden, Kern, Ungar, and
  Eichstaedt}]{Guntuku2017DetectingDA}
Sharath~Chandra Guntuku, David~Bryce Yaden, Margaret~L. Kern, Lyle~H. Ungar,
  and Johannes~C. Eichstaedt. 2017.
\newblock \href {https://api.semanticscholar.org/CorpusID:53273218} {Detecting
  depression and mental illness on social media: an integrative review}.
\newblock \emph{Current Opinion in Behavioral Sciences}, 18:43--49.

\bibitem[{Hipson and Mohammad(2021)}]{hipson2021emotion}
Will~E. Hipson and Saif~M. Mohammad. 2021.
\newblock \href {https://doi.org/10.1371/journal.pone.0256153} {Emotion
  dynamics in movie dialogues}.
\newblock \emph{PLOS ONE}, 16:1--19.

\bibitem[{Hirschfeld(2001)}]{PMID:15014592}
Robert M.~A. Hirschfeld. 2001.
\newblock \href {https://doi.org/10.4088/pcc.v03n0609} {The comorbidity of
  major depression and anxiety disorders: Recognition and management in primary
  care}.
\newblock \emph{Primary care companion to the Journal of clinical psychiatry},
  3(6):244—254.

\bibitem[{Hoemann et~al.(2021{\natexlab{a}})Hoemann, Feldman~Barrett, and
  Quigley}]{emo_gran_icc}
Katie Hoemann, Lisa Feldman~Barrett, and Karen~S. Quigley. 2021{\natexlab{a}}.
\newblock \href {https://doi.org/10.3389/fpsyg.2021.704125} {Emotional
  granularity increases with intensive ambulatory assessment: Methodological
  and individual factors influence how much}.
\newblock \emph{Frontiers in Psychology}, 12.

\bibitem[{Hoemann et~al.(2021{\natexlab{b}})Hoemann, Nielson, Yuen, Gurera,
  Quigley, and Barrett}]{Hoemann2021ExpertiseIE}
Katie Hoemann, Cathy Nielson, Ashley Yuen, Jacob Gurera, Karen~S. Quigley, and
  Lisa~Feldman Barrett. 2021{\natexlab{b}}.
\newblock \href {https://api.semanticscholar.org/CorpusID:247220000} {Expertise
  in emotion: A scoping review and unifying framework for individual
  differences in the mental representation of emotional experience.}
\newblock \emph{Psychological bulletin}, 147 11:1159--1183.

\bibitem[{Houben et~al.(2015)Houben, Van Den~Noortgate, and
  Kuppens}]{Houben2015}
Marlies Houben, Wim Van Den~Noortgate, and Peter Kuppens. 2015.
\newblock The relation between short-term emotion dynamics and psychological
  well-being: A meta-analysis.

\bibitem[{Kashdan et~al.(2015)Kashdan, Barrett, and
  McKnight}]{kashdan2015unpacking}
Todd~B Kashdan, Lisa~Feldman Barrett, and Patrick~E McKnight. 2015.
\newblock Unpacking emotion differentiation: Transforming unpleasant experience
  by perceiving distinctions in negativity.
\newblock \emph{Current Directions in Psychological Science}, 24(1):10--16.

\bibitem[{Koops et~al.(2023)Koops, Brederoo, de~Boer, Nadema, Voppel, and
  Sommer}]{koops}
Sanne Koops, Sanne~G Brederoo, Janna~N de~Boer, Femke~G Nadema, Alban~E Voppel,
  and Iris~E Sommer. 2023.
\newblock \href {https://doi.org/10.2174/1871527320666211213125847} {Speech as
  a biomarker for depression}.
\newblock \emph{CNS\&; neurological disorders drug targets}, 22(2):152—160.

\bibitem[{Kragel et~al.(2022)Kragel, Hariri, and LaBar}]{Kragel2022}
Philip~A. Kragel, Ahmad~R. Hariri, and Kevin~S. LaBar. 2022.
\newblock \href {https://doi.org/10.1162/jocn_a_01787} {The temporal dynamics
  of spontaneous emotional brain states and their implications for mental
  health}.
\newblock \emph{Journal of cognitive neuroscience}, 34(5):715--728.
\newblock May, 2022.

\bibitem[{Kring et~al.(2003)Kring, Barrett, and Gard}]{kring2003broad}
Ann~M Kring, Lisa~Feldman Barrett, and David~E Gard. 2003.
\newblock On the broad applicability of the affective circumplex:
  representations of affective knowledge among schizophrenia patients.
\newblock \emph{Psychological Science}, 14(3):207--214.

\bibitem[{Kuppens and Verduyn(2017)}]{KUPPENS201722}
Peter Kuppens and Philippe Verduyn. 2017.
\newblock \href {https://doi.org/https://doi.org/10.1016/j.copsyc.2017.06.004}
  {Emotion dynamics}.
\newblock \emph{Current Opinion in Psychology}, 17:22--26.
\newblock Emotion.

\bibitem[{Latkin et~al.(2017)Latkin, Edwards, Davey-Rothwell, and
  Tobin}]{latkin2017relationship}
Carl~A Latkin, Catie Edwards, Melissa~A Davey-Rothwell, and Karin~E Tobin.
  2017.
\newblock The relationship between social desirability bias and self-reports of
  health, substance use, and social network factors among urban substance users
  in baltimore, maryland.
\newblock \emph{Addictive behaviors}, 73:133--136.

\bibitem[{Lee et~al.(2017)Lee, Lindquist, and Nam}]{emo_gran_brain}
Ja~Y. Lee, Kristen~A. Lindquist, and Chang~S. Nam. 2017.
\newblock \href {https://doi.org/10.3389/fnhum.2017.00133} {Emotional
  granularity effects on event-related brain potentials during affective
  picture processing}.
\newblock \emph{Frontiers in Human Neuroscience}, 11.

\bibitem[{Lena(2021)}]{biosocial}
Palaniyappan Lena. 2021.
\newblock \href
  {https://www.proquest.com/scholarly-journals/more-than-biomarker-could-language-be-biosocial/docview/2567801968/se-2}
  {More than a biomarker: could language be a biosocial marker of psychosis?}
\newblock \emph{NPJ Schizophrenia}, 7(1).
\newblock Copyright - © The Author(s) 2021. This work is published under
  http://creativecommons.org/licenses/by/4.0/ (the “License”).
  Notwithstanding the ProQuest Terms and Conditions, you may use this content
  in accordance with the terms of the License; Last updated - 2023-02-22.

\bibitem[{Lewis et~al.(2010)Lewis, Haviland-Jones, and
  Barrett}]{lewis2010handbook}
Michael Lewis, Jeannette~M Haviland-Jones, and Lisa~Feldman Barrett. 2010.
\newblock \emph{Handbook of emotions}.
\newblock Guilford Press.

\bibitem[{Lindquist and Barrett(2008)}]{lindquist2008emotional}
Kristen~A Lindquist and Lisa~Feldman Barrett. 2008.
\newblock Emotional complexity.
\newblock \emph{Handbook of emotions}, 4:513--530.

\bibitem[{Losada et~al.(2017)Losada, Crestani, and Parapar}]{eRisk2017}
David~E Losada, Fabio Crestani, and Javier Parapar. 2017.
\newblock Clef 2017 erisk overview: Early risk prediction on the internet:
  Experimental foundations.
\newblock pages 346--360.

\bibitem[{Losada et~al.(2018)Losada, Crestani, and Parapar}]{eRisk2018}
David~E Losada, Fabio Crestani, and Javier Parapar. 2018.
\newblock Overview of erisk: early risk prediction on the internet.
\newblock In \emph{Experimental IR Meets Multilinguality, Multimodality, and
  Interaction: 9th International Conference of the CLEF Association, CLEF 2018,
  Avignon, France, September 10-14, 2018, Proceedings 9}, pages 343--361.
  Springer.

\bibitem[{Miller(1995)}]{10.1145/219717.219748}
George~A. Miller. 1995.
\newblock \href {https://doi.org/10.1145/219717.219748} {Wordnet: a lexical
  database for english}.
\newblock \emph{Commun. ACM}, 38(11):39–41.

\bibitem[{Mohammad(2023)}]{mohammad-2023-best}
Saif Mohammad. 2023.
\newblock \href {https://doi.org/10.18653/v1/2023.findings-eacl.136} {Best
  practices in the creation and use of emotion lexicons}.
\newblock In \emph{Findings of the Association for Computational Linguistics:
  EACL 2023}, pages 1825--1836, Dubrovnik, Croatia. Association for
  Computational Linguistics.

\bibitem[{Mohammad(2018)}]{LREC18-AIL}
Saif~M. Mohammad. 2018.
\newblock Word affect intensities.
\newblock In \emph{Proceedings of the 11th Edition of the Language Resources
  and Evaluation Conference (LREC-2018)}, Miyazaki, Japan.

\bibitem[{Mohammad(2022)}]{mohammad-2022-ethics-sheet}
Saif~M. Mohammad. 2022.
\newblock \href {https://doi.org/10.1162/coli_a_00433} {Ethics sheet for
  automatic emotion recognition and sentiment analysis}.
\newblock \emph{Computational Linguistics}, 48(2):239--278.

\bibitem[{Pollack(2005)}]{pollack2005comorbid}
Mark~H Pollack. 2005.
\newblock Comorbid anxiety and depression.
\newblock \emph{Journal of Clinical Psychiatry}, 66:22.

\bibitem[{Pond~Jr et~al.(2012)Pond~Jr, Kashdan, DeWall, Savostyanova, Lambert,
  and Fincham}]{pond2012emotion}
Richard~S Pond~Jr, Todd~B Kashdan, C~Nathan DeWall, Antonina Savostyanova,
  Nathaniel~M Lambert, and Frank~D Fincham. 2012.
\newblock Emotion differentiation moderates aggressive tendencies in angry
  people: A daily diary analysis.
\newblock \emph{Emotion}, 12(2):326.

\bibitem[{Reitsema et~al.(2022)Reitsema, Jeronimus, van Dijk, and
  de~Jonge}]{reitsema2022emotion}
Anne~M Reitsema, Bertus~F Jeronimus, Marijn van Dijk, and Peter de~Jonge. 2022.
\newblock Emotion dynamics in children and adolescents: A meta-analytic and
  descriptive review.
\newblock \emph{Emotion}, 22(2):374.

\bibitem[{Resnik(1995)}]{Resnik1995UsingIC}
Philip Resnik. 1995.
\newblock \href {https://api.semanticscholar.org/CorpusID:1752785} {Using
  information content to evaluate semantic similarity in a taxonomy}.
\newblock In \emph{International Joint Conference on Artificial Intelligence}.

\bibitem[{Seabold and Perktold(2010)}]{seabold2010statsmodels}
Skipper Seabold and Josef Perktold. 2010.
\newblock statsmodels: Econometric and statistical modeling with python.
\newblock In \emph{9th Python in Science Conference}.

\bibitem[{Seabrook et~al.(2018)Seabrook, Kern, Fulcher, and Rickard}]{seabrook}
Elizabeth~M Seabrook, Margaret~L Kern, Ben~D Fulcher, and Nikki~S Rickard.
  2018.
\newblock \href {https://doi.org/10.2196/jmir.9267} {Predicting depression from
  language-based emotion dynamics: Longitudinal analysis of facebook and
  twitter status updates}.
\newblock \emph{J Med Internet Res}, 20(5):e168.

\bibitem[{Seah et~al.(2020)Seah, Aurora, and Coifman}]{seah2020emotion}
TH~Stanley Seah, Pallavi Aurora, and Karin~G Coifman. 2020.
\newblock Emotion differentiation as a protective factor against the behavioral
  consequences of rumination: A conceptual replication and extension in the
  context of social anxiety.
\newblock \emph{Behavior Therapy}, 51(1):135--148.

\bibitem[{Shrout and Fleiss(1979)}]{shrout1979intraclass}
Patrick~E Shrout and Joseph~L Fleiss. 1979.
\newblock Intraclass correlations: uses in assessing rater reliability.
\newblock \emph{Psychological bulletin}, 86(2):420.

\bibitem[{Silk et~al.(2011)Silk, Forbes, Whalen, Jakubcak, Thompson, Ryan,
  Axelson, Birmaher, and Dahl}]{depressedYouth}
Jennifer~S. Silk, Erika~E. Forbes, Diana~J. Whalen, Jennifer~L. Jakubcak,
  Wesley~K. Thompson, Neal~D. Ryan, David~A. Axelson, Boris Birmaher, and
  Ronald~E. Dahl. 2011.
\newblock \href {https://doi.org/https://doi.org/10.1016/j.jecp.2010.10.007}
  {Daily emotional dynamics in depressed youth: A cell phone ecological
  momentary assessment study}.
\newblock \emph{Journal of Experimental Child Psychology}, 110(2):241--257.
\newblock Special Issue: Assessment of Emotion in Children and Adolescents.

\bibitem[{Sperry et~al.(2020)Sperry, Walsh, and Kwapil}]{sperry}
Sarah~H. Sperry, Molly~A. Walsh, and Thomas~R. Kwapil. 2020.
\newblock \href {https://doi.org/https://doi.org/10.1016/j.jad.2019.09.076}
  {Emotion dynamics concurrently and prospectively predict mood
  psychopathology}.
\newblock \emph{Journal of Affective Disorders}, 261:67--75.

\bibitem[{Starr et~al.(2017)Starr, Hershenberg, Li, and
  Shaw}]{starr2017feelings}
Lisa~R Starr, Rachel Hershenberg, Y~Irina Li, and Zoey~A Shaw. 2017.
\newblock When feelings lack precision: Low positive and negative emotion
  differentiation and depressive symptoms in daily life.
\newblock \emph{Clinical Psychological Science}, 5(4):613--631.

\bibitem[{Suhavi et~al.(2022)Suhavi, Singh, Arora, Shrivastava, Singh, Shah,
  and Kumaraguru}]{_Singh_Arora_Shrivastava_Singh_Shah_Kumaraguru_2022}
Suhavi, Asmit~Kumar Singh, Udit Arora, Somyadeep Shrivastava, Aryaveer Singh,
  Rajiv~Ratn Shah, and Ponnurangam Kumaraguru. 2022.
\newblock \href {https://doi.org/10.1609/icwsm.v16i1.19368} {Twitter-stmhd: An
  extensive user-level database of multiple mental health disorders}.
\newblock \emph{Proceedings of the International AAAI Conference on Web and
  Social Media}, 16(1):1182--1191.

\bibitem[{Suvak et~al.(2011)Suvak, Litz, Sloan, Zanarini, Barrett, and
  Hofmann}]{suvak2011emotional}
Michael~K Suvak, Brett~T Litz, Denise~M Sloan, Mary~C Zanarini, Lisa~Feldman
  Barrett, and Stefan~G Hofmann. 2011.
\newblock Emotional granularity and borderline personality disorder.
\newblock \emph{Journal of abnormal psychology}, 120(2):414.

\bibitem[{Tedeschi(2013)}]{tedeschi2013impression}
James~T Tedeschi. 2013.
\newblock \emph{Impression management theory and social psychological
  research}.
\newblock Academic Press.

\bibitem[{Teodorescu et~al.(2023)Teodorescu, Cheng, Fyshe, and
  Mohammad}]{teodorescu-etal-2023-language}
Daniela Teodorescu, Tiffany Cheng, Alona Fyshe, and Saif Mohammad. 2023.
\newblock \href {https://doi.org/10.18653/v1/2023.emnlp-main.188} {Language and
  mental health: Measures of emotion dynamics from text as linguistic biosocial
  markers}.
\newblock In \emph{Proceedings of the 2023 Conference on Empirical Methods in
  Natural Language Processing}, pages 3117--3133, Singapore. Association for
  Computational Linguistics.

\bibitem[{Teodorescu and Mohammad(2023)}]{teodorescu2023evaluating}
Daniela Teodorescu and Saif Mohammad. 2023.
\newblock Evaluating emotion arcs across languages: Bridging the global divide
  in sentiment analysis.
\newblock In \emph{Findings of the Association for Computational Linguistics:
  EMNLP 2023}, pages 4124--4137.

\bibitem[{Tugade et~al.(2004)Tugade, Fredrickson, and
  Barrett}]{Tugade2004PsychologicalRA}
Michele~M. Tugade, Barbara~L. Fredrickson, and Lisa~Feldman Barrett. 2004.
\newblock \href {https://api.semanticscholar.org/CorpusID:16323336}
  {Psychological resilience and positive emotional granularity: examining the
  benefits of positive emotions on coping and health.}
\newblock \emph{Journal of personality}, 72 6:1161--90.

\bibitem[{Virtanen et~al.(2020)Virtanen, Gommers, Oliphant, Haberland, Reddy,
  Cournapeau, Burovski, Peterson, Weckesser, Bright, {van der Walt}, Brett,
  Wilson, Millman, Mayorov, Nelson, Jones, Kern, Larson, Carey, Polat, Feng,
  Moore, {VanderPlas}, Laxalde, Perktold, Cimrman, Henriksen, Quintero, Harris,
  Archibald, Ribeiro, Pedregosa, {van Mulbregt}, and {SciPy 1.0
  Contributors}}]{2020SciPy-NMeth}
Pauli Virtanen, Ralf Gommers, Travis~E. Oliphant, Matt Haberland, Tyler Reddy,
  David Cournapeau, Evgeni Burovski, Pearu Peterson, Warren Weckesser, Jonathan
  Bright, St{\'e}fan~J. {van der Walt}, Matthew Brett, Joshua Wilson, K.~Jarrod
  Millman, Nikolay Mayorov, Andrew R.~J. Nelson, Eric Jones, Robert Kern, Eric
  Larson, C~J Carey, {\.I}lhan Polat, Yu~Feng, Eric~W. Moore, Jake
  {VanderPlas}, Denis Laxalde, Josef Perktold, Robert Cimrman, Ian Henriksen,
  E.~A. Quintero, Charles~R. Harris, Anne~M. Archibald, Ant{\^o}nio~H. Ribeiro,
  Fabian Pedregosa, Paul {van Mulbregt}, and {SciPy 1.0 Contributors}. 2020.
\newblock \href {https://doi.org/10.1038/s41592-019-0686-2} {{{SciPy} 1.0:
  Fundamental Algorithms for Scientific Computing in Python}}.
\newblock \emph{Nature Methods}, 17:261--272.

\bibitem[{Vishnubhotla and Mohammad(2022)}]{vishnubhotla-mohammad-2022-tusc}
Krishnapriya Vishnubhotla and Saif~M. Mohammad. 2022.
\newblock \href {https://aclanthology.org/2022.lrec-1.442} {{Tweet Emotion
  Dynamics}: Emotion word usage in tweets from {US} and {C}anada}.
\newblock In \emph{Proceedings of the Thirteenth Language Resources and
  Evaluation Conference}, pages 4162--4176, Marseille, France. European
  Language Resources Association.

\bibitem[{Willroth et~al.(2020)Willroth, Flett, and
  Mauss}]{willroth2020depressive}
Emily~C Willroth, Jayde~AM Flett, and Iris~B Mauss. 2020.
\newblock Depressive symptoms and deficits in stress-reactive negative,
  positive, and within-emotion-category differentiation: A daily diary study.
\newblock \emph{Journal of personality}, 88(2):174--184.

\end{thebibliography}

\appendix
\section{Twitter-STMHD Dataset}
\label{app:twitter_dataset}
\citet{_Singh_Arora_Shrivastava_Singh_Shah_Kumaraguru_2022} created a regular expression pattern to identify posts which contained a self-disclosure of a diagnosis and the diagnosis name (using a lexicon of common synonyms, abbreviations, etc.) such as `diagnosed with X'. 
They collected a large set of tweets using the regex. 
This resulted in a preliminary dataset of users with potential MHC diagnoses.
To handle false positives (e.g., `my family member has been diagnosed with X', or `I was not diagnosed with X'), the dataset was split into two non-overlapping parts, one of which was manually annotated, and the other using an updated and high-precision regex.
In the part that was annotated by hand, each tweet was annotated by two members of the team. A user was only included in the dataset if both annotations were positive as self-disclosing for a particular class. A licensed clinical psychologist found the 500-tweet sample to be 99.2\% accurate. 
The manual annotations were used to refine the regular expressions and diagnosis name lexicon.
This updated search pattern was applied to the other dataset split.
To verify the quality of the updated regex, the authors applied it to the manually annotated dataset split.
When considering the manual annotations as correct, the regex was found to be 94\% accurate.

\section{Lexicon Words Removed}
\label{app:lex_words_removed}
We considered the following sets of terms to be stop-words, which do not contribute to the emotion score of an utterance, for our analysis:
\begin{compactitem}
    \item \textbf{Common stopwords:} We remove common English stopwords, such as \textit{the}, \textit{of}, \textit{for}, etc. We use the list of English stopwords from the Python NLTK library. The full list can be found at \url{https://gist.github.com/sebleier/554280}.
    \item \textbf{Domain-specific stopwords: }We remove terms (words and word pairs) whose dominant usage on social media platforms differs from their annotated sense (e.g, \textit{like}, \textit{chaotic evil}, \textit{good morning}). The full list of these terms is in Table \ref{tab:gen_list}.
    \item \textbf{MHC-associated terms: } Finally, we filter out terms that are explicitly associated with the MHCs that we consider, such as \textit{anxiety}, \textit{mental health}, and \textit{panic attack}. The full list of terms is in Table \ref{tab:zero_freqs_list}.
\end{compactitem}
\begin{table*}[t]
\centering
{\small
\begin{tabular}{llll}
 \hline
love & flu shot & raptor  & discord \\
christmas & good day & good morning & good evening\\
birthday  & good night & good afternoon & bloody murder\\
pretty  & true crime & full time & gut punch\\
vibe  & wholesome content & slur word & life time\\
vote  &jump scares & hot chocolate & chaotic evil\\
trump  & fever dream & chaotic energy & chaotic good\\
like & guilty pleasure & chaotic neutral & hot mess\\

\hline 
\end{tabular}
}
\caption{Twitter-specific words and bigrams removed from the emotion lexicons.}
\label{tab:gen_list}
\end{table*}

\begin{table*}[t]
\centering
{\small
\begin{tabular}{ll}
\hline
disability & ptsd \\
psychosis & adhd \\
suicide & depressive \\
depressed & disorder \\
anxiety & mental health \\
anxious & mental illness \\
disabled & panic attack \\
\hline 
\end{tabular}
}
\caption{Mental health specific terms removed from the emotion lexicons.}
\label{tab:zero_freqs_list}
\end{table*}

\section{Statistical Assumptions}
\label{app:statistical_assumptions}

Below we describe in more depth the requirements for performing an independent t-test, which was done in our analyses. 
\begin{compactitem}
    \item \textbf{The dependent variable must be measured using a continuous scale}: emotion granularity is measured as the average of Spearman correlation between emotion arcs in the group, resulting in continuous values. 
    \item \textbf{The independent variable must have two categorical and independent groups}: Our independent variable is diagnosis, which is either an MHC or the control group.
    \item \textbf{Independence of observations}: Since the text stream of utterances come from different people, we can assume these are independent observations.\footnote{In reality individuals are largely influenced by one another as we see, interact and engage with content from various communities, which can influence our emotional state and therefore utterances. However, for the purposes of our experiments since the utterances come from different people we can assume they are independent.}
    \item \textbf{Approximately normally distributed dependent variable for each group of independent variable}: Given the large number of people and number of utterances per person in our dataset, we can assume that the means of the data for each group is approximately normally distributed according to the law of large numbers. Further, the t-test is robust to violations of normality.
    \item \textbf{Homogeneity of variance}: We performed Levene's test for homogeneity of variance to verify whether this assumption is met. Our data did not meet this assumption, therefore we performed t-tests with the unequal variance setting as True in \texttt{SciPy}.
\end{compactitem}

\section{Emotion Lexicons}
\label{app:emo-lex}
In Table \ref{tab:emo-lex-counts}, we report statistics on the number of emotion terms in each lexicon for the eight emotions we consider in this work, and the number of terms common to and mutually-exclusive between each emotion pair.

\begin{table*}
\centering
{\small
\begin{tabular}{llrrrrr}
\toprule
\textbf{emo1} & \textbf{emo2} & \textbf{e1-all} & \textbf{e2-all} & \textbf{e12-comm} & \textbf{e1-excl} & \textbf{e2-excl} \\
\midrule
anger & anticipation & 1157 & 782 & 43 & 1114 & 739 \\
anger & disgust & 1157 & 886 & 407 & 750 & 479 \\
anger & fear & 1157 & 1343 & 551 & 606 & 792 \\
anger & joy & 1157 & 946 & 3 & 1154 & 943 \\
anger & sadness & 1157 & 1014 & 382 & 775 & 632 \\
anger & surprise & 1157 & 454 & 102 & 1055 & 352 \\
anger & trust & 1157 & 1332 & 6 & 1151 & 1326 \\
anticipation & disgust & 782 & 886 & 19 & 763 & 867 \\
anticipation & fear & 782 & 1343 & 82 & 700 & 1261 \\
anticipation & joy & 782 & 946 & 283 & 499 & 663 \\
anticipation & sadness & 782 & 1014 & 32 & 750 & 982 \\
anticipation & surprise & 782 & 454 & 131 & 651 & 323 \\
anticipation & trust & 782 & 1332 & 283 & 499 & 1049 \\
disgust & fear & 886 & 1343 & 400 & 486 & 943 \\
disgust & joy & 886 & 946 & 1 & 885 & 945 \\
disgust & sadness & 886 & 1014 & 336 & 550 & 678 \\
disgust & surprise & 886 & 454 & 56 & 830 & 398 \\
disgust & trust & 886 & 1332 & 2 & 884 & 1330 \\
fear & joy & 1343 & 946 & 2 & 1341 & 944 \\
fear & sadness & 1343 & 1014 & 545 & 798 & 469 \\
fear & surprise & 1343 & 454 & 137 & 1206 & 317 \\
fear & trust & 1343 & 1332 & 9 & 1334 & 1323 \\
joy & sadness & 946 & 1014 & 0 & 946 & 1014 \\
joy & surprise & 946 & 454 & 113 & 833 & 341 \\
joy & trust & 946 & 1332 & 308 & 638 & 1024 \\
sadness & surprise & 1014 & 454 & 73 & 941 & 381 \\
sadness & trust & 1014 & 1332 & 3 & 1011 & 1329 \\
surprise & trust & 454 & 1332 & 56 & 398 & 1276 \\
\bottomrule
\end{tabular}}
\caption{\textbf{Emotion Lexicons: }
For each emotion pair (emo1, emo2), the number of terms in each lexicon (e1-all, e2-all), the number of emotion terms common to the two lexicons (e12-comm), and the number of mutually-exclusive emotion terms (e1-excl, e2-excl).}
\label{tab:emo-lex-counts}
\end{table*}

\section{Visualization of Emotion Arcs}
\label{app:viz}
\begin{figure*}
    \centering
    \includegraphics[scale=0.45, width=\textwidth]{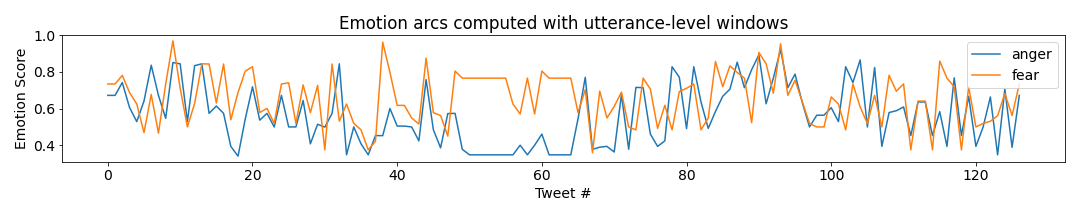}
    \caption{\textbf{Emotion arcs: }
    Tweet-level emotion arcs for anger and fear, for a sampled user from the Twitter-STMHD dataset.}
    \label{fig:sample-arcs}
\end{figure*}

In Figure \ref{fig:sample-arcs}, we plot the emotion arcs for the anger-fear emotion pair, from the negative valence group, for a tweeter from an MHC group of the Twitter-STMHD dataset. Emotion scores are computed and plotted at the utterance-level, i.e, independently for each tweet by the user. Note that larger window sizes and overlapping windows will lead to smoother arcs.

\section{Emotion Granularity: Hyperparameters}
\label{app:hyper1}
We report the results of the statistical analyses of emotion granularity when emotion arcs were generated using different choices of the hyperparameters described in Section \ref{sec-arcs}. 
Table \ref{tab:arrows_all_emotions_no_comorbid_zero-0-0_ex} reports the results when non-lexicon terms (and tweets) are assigned a score of 0, and only mutually-exclusive emotion terms are considered, similar to Table \ref{tab:arrows_all_emotions_no_comorbid_zero-50-25_ex}, but no user thresholds on number of tweets and unique emotion terms are applied. Table \ref{tab:arrows_all_emotions_no_comorbid_nan-50-25_ex} reports the results when non-lexicon terms (and tweets) are not considered, and user thresholds are set to 50 and 25, (similar to Table \ref{tab:arrows_all_emotions_no_comorbid_zero-50-25_ex}), and only mutually-exclusive emotion terms are considered (similar to Table \ref{tab:arrows_all_emotions_no_comorbid_zero-50-25_ex}).

We find that largely the results do not change. However, when non-lexicon terms and tweets are ignored, this results in a smaller set of tweets to compute the emotion arc over, and fewer tweeters who meet the user thresholds for each group. This results in signals turning off for certain MHCs. 

\begin{table*}[t]
\centering
{\small
  \begin{tabular}{lcccccccc}
\hline
\textbf{Dataset, MHC--Control} &
 $IC(n)$ & $IC(v)$ &
 $EG(pos)$ & 
  $EG(neg)$ &
  $EG(var)$ &
  $EG(cross)$ &
  $EG(overall)$ \\
 \hline
\textit{Twitter-STMHD} \\
$\;\;\;$ ADHD--control & -- & -- & lower & lower & lower & lower & lower \\  
$\;\;\;$ Anxiety--control & -- & -- & lower & lower & lower & lower & lower \\  
$\;\;\;$ Bipolar--control & -- & -- & -- & lower & lower & lower & lower \\     
$\;\;\;$ MDD--control & -- & -- & lower & -- & -- & lower & lower \\    
$\;\;\;$ OCD--control & -- & --  & lower & lower & lower & lower & lower \\    
$\;\;\;$ PPD--control & -- & -- & -- & lower & lower & -- & lower \\    
$\;\;\;$ PTSD--control & -- & -- & lower & lower & lower & lower & lower \\     
$\;\;\;$ Depression--control & -- & -- & lower & lower & lower & lower & lower\\     
\textit{Reddit eRisk} \\
$\;\;\;$ Depression--control & -- & -- & lower & lower & lower & lower & lower \\ 
\hline 
\end{tabular}
}
\caption{\textbf{Emotion Granularity - hyperparameter variations}: 
The difference in emotion granularity between each MHC group and the control. A significant difference is indicated by the word `lower' or `higher', indicating the direction of the difference in granularity. Non-lexicon terms and tweets are assigned a score of zero; user tweet and unique term thresholds are both set to 0, and only mutually-exclusive emotion terms are considered.
}
\label{tab:arrows_all_emotions_no_comorbid_zero-0-0_ex}
\end{table*}

\begin{table*}[t]
\centering
{\small
  \begin{tabular}{lcccccccc}
\hline
\textbf{Dataset, MHC--Control} &
 $IC(n)$ & $IC(v)$ &
 $EG(pos)$ & 
  $EG(neg)$ &
  $EG(var)$ &
  $EG(cross)$ &
  $EG(overall)$ \\
 \hline
\textit{Twitter-STMHD} \\
$\;\;\;$ ADHD--control & -- & -- & lower & lower & lower & lower & lower \\ 
$\;\;\;$ Anxiety--control & -- & -- & -- & lower & lower & lower & lower \\ 
$\;\;\;$ Bipolar--control & -- & -- & lower & lower & lower & lower & lower \\    
$\;\;\;$ MDD--control & -- & -- & -- & -- & -- & -- & -- \\   
$\;\;\;$ OCD--control & -- & -- & lower & lower & lower & lower & lower \\   
$\;\;\;$ PPD--control & -- & -- & lower & lower & lower & lower & lower \\   
$\;\;\;$ PTSD--control & -- & -- & lower & lower & lower & lower & lower \\    
$\;\;\;$ Depression--control & -- & -- & higher & lower & lower & lower & -- \\    
\textit{Reddit eRisk} \\
$\;\;\;$ Depression--control & -- & -- & -- & lower & lower & lower & lower \\ 
\hline 
\end{tabular}
}
\caption{\textbf{Emotion Granularity - hyperparameter variations}: 
 The difference in emotion granularity between each MHC group and the control. A significant difference is indicated by the word `lower' or `higher', indicating the direction of the difference in granularity. Non-lexicon terms and tweets are discarded; user tweet and unique term thresholds are set to 50 and 25, and only mutually-exclusive emotion terms are considered.
}
 \vspace*{-3mm}
\label{tab:arrows_all_emotions_no_comorbid_nan-50-25_ex}
\end{table*}

\subsection{Various Window Sizes}
\label{app:emo_granularity_diff_window_sizes}
We report the results of the statistical analyses of emotion granularity when emotion arcs were generated using two other window sizes: 100 (Table \ref{tab:arrows_all_emotions_no_comorbid_window_100_gran}) and 500 (Table \ref{tab:arrows_all_emotions_no_comorbid_window_500}). All other hyperparameters are the same as for Table \ref{tab:arrows_all_emotions_no_comorbid_zero-50-25_ex}.

We find that largely the results do not change, however there are some differences in the scenario when the dataset was smaller (e.g., eRisk dataset or MHC such as MDD). In such cases, when the window size is increased, it is possible that several emotional experiences occurred, resulting in a weaker signal of emotion granularity.

\begin{table*}[t]
\centering
{\small
\begin{tabular}{lllllllll}
\hline
\textbf{Dataset, MHC--Control} &
$IC(n)$ & $IC(v)$ &
  $EG(pos)$ &
  $EG(neg)$ &
  $EG(var)$ &
  $EG(cross)$ &
  $EG(overall)$ \\
 \hline
\textit{Twitter-STMHD} \\
$\;\;\;$ ADHD--control & -- & -- & lower & lower & higher & lower & lower \\ 
$\;\;\;$ Anxiety--control & -- & -- & lower & lower & lower & higher & lower \\
$\;\;\;$ Bipolar--control & -- & -- & lower & lower & lower & -- & lower \\ 
$\;\;\;$ MDD--control & -- & -- & lower & lower & -- & -- & lower \\ 
$\;\;\;$ OCD--control & -- & -- & lower & lower & -- & -- & lower \\ 
$\;\;\;$ PPD--control & -- & -- & -- & -- & lower & higher & -- \\ 
$\;\;\;$ PTSD--control & -- & -- & -- & lower & -- & higher & lower \\ 
$\;\;\;$ Depression--control & -- & -- & lower & lower & lower & higher & lower \\ 
\textit{Reddit eRisk} \\
$\;\;\;$ Depression--control & -- & -- & lower & -- & lower & -- & lower \\ 
\hline 
\end{tabular}
}
\caption{ \textbf{Emotion Granularity - using window 100}: 
The difference in emotion granularity between each MHC group and the control. A significant difference is indicated by the word `lower' or `higher', indicating the direction of the difference in granularity.
}
\label{tab:arrows_all_emotions_no_comorbid_window_100_gran}
\end{table*}

\begin{table*}[t]
\centering
{\small
\begin{tabular}{lllllllll}
\hline
\textbf{Dataset, MHC--Control} &
  $IC(n)$ & $IC(v)$ &
  $EG(pos)$ &
  $EG(neg)$ &
  $EG(var)$ &
  $EG(cross)$ &
  $EG(overall)$ \\
 \hline
\textit{Twitter-STMHD} \\
$\;\;\;$ ADHD--control & -- & -- & lower & lower & -- & lower & lower\\ 
$\;\;\;$ Anxiety--control & -- & -- & lower & lower & lower & higher  & lower\\
$\;\;\;$ Bipolar--control & -- & -- & lower & lower & lower & -- & lower\\ 
$\;\;\;$ MDD--control & -- & -- & -- & -- & -- & -- & lower\\ 
$\;\;\;$ OCD--control & -- & -- & lower & -- & -- & -- & --\\ 
$\;\;\;$ PPD--control & -- & -- & -- & -- & -- & higher & --\\ 
$\;\;\;$ PTSD--control & -- & -- & -- & lower & -- & higher & lower\\ 
$\;\;\;$ Depression--control & -- & -- & lower & lower & lower & higher & lower\\ 
\textit{Reddit eRisk} \\
$\;\;\;$ Depression--control & -- & -- & lower & -- & -- & -- & --\\ 
\hline 
\end{tabular}
}
\caption{ \textbf{Emotion Granularity - using window 500}: 
The difference in emotion granularity between each MHC group and the control. A significant difference is indicated by the word `lower' or `higher', indicating the direction of the difference in granularity.
}
\label{tab:arrows_all_emotions_no_comorbid_window_500}
\end{table*}

\section{Emotion Granularity: Emotion Pairs}
\label{app:emo_gran_emo_pairs}
In Table \ref{tab:arrows_all_no_comorbid_v2} 
we report the pairwise emotion granularity results when testing for significant differences between MHCs and the control group.

\begin{table*}[t]
\centering
{\small
\begin{tabular}{lllllllll}
\hline
\textbf{Emotion Pair, MHC-Control } & \textbf{ADHD} & \textbf{Anxiety} & \textbf{Bipolar} & \textbf{Depression} & \textbf{MDD} & \textbf{OCD} & \textbf{PPD} & \textbf{PTSD} \\  \hline

anger--anticipation & lower & lower & -- & lower & -- & lower & -- & lower \\
anger--disgust & lower & lower & lower & lower & lower & lower & -- & lower \\
anger--fear & lower & lower & lower & lower & -- & lower & lower & lower \\
anger--joy & lower & lower & higher & lower & -- & lower & -- & -- \\
anger--sadness & lower & lower & lower & lower & -- & lower & lower & lower \\
anger--surprise & lower & lower & -- & lower & -- & lower & -- & lower \\
anger--trust & lower & lower & lower & lower & lower & lower & -- & lower \\ \hline

anticipation--disgust & lower & lower & lower & lower & -- & lower & -- & lower \\
anticipation--fear & lower & lower & -- & lower & lower & lower & lower & lower \\
anticipation--joy & lower & lower & lower & lower & lower & lower & lower & lower \\
anticipation--sadness & lower & lower & lower & lower & -- & lower & -- & lower \\
anticipation--surprise & lower & lower & lower & lower & -- & lower & -- & lower \\
anticipation--trust & lower & lower & lower & lower & lower & lower & lower & lower \\ \hline

disgust--fear & lower & lower & lower & lower & -- & lower & lower & lower \\
disgust--joy & lower & lower & -- & lower & -- & lower & -- & lower \\
disgust--sadness & lower & lower & lower & lower & -- & lower & lower & lower \\
disgust--surprise & lower & lower & lower & lower & -- & lower & -- & lower \\
disgust--trust & lower & lower & lower & lower & lower & lower & lower & lower \\ \hline

fear--joy & lower & lower & higher & lower & -- & lower & -- & lower \\
fear--sadness & lower & lower & lower & lower & -- & lower & lower & lower \\
fear--surprise & lower & lower & lower & lower & -- & lower & -- & lower \\
fear--trust & lower & lower & lower & lower & lower & lower & lower & lower \\ \hline

joy--sadness & lower & lower & -- & lower & -- & lower & -- & lower \\
joy--surprise & lower & lower & -- & lower & -- & lower & -- & lower \\
joy--trust & lower & lower & lower & lower & lower & lower & -- & lower \\ \hline

sadness--surprise & lower & lower & lower & lower & -- & lower & lower & lower \\
sadness--trust & lower & lower & lower & lower & lower & lower & lower & lower \\ \hline

surprise--trust & lower & lower & -- & lower & -- & lower & lower & lower \\
\hline 
\end{tabular}
}
\caption{ \textbf{Emotion Granularity - emotion Pairs}: 
The difference in emotion granularity between each emotion pair, for each MHC group and the control in the Twitter-STMHD dataset. A significant difference is indicated by the word `lower' or `higher', indicating the direction of the difference.
}
\label{tab:arrows_all_no_comorbid_v2}
\end{table*}

\section{Term Specificity Results}
Table 
\label{app:word_specificity}
\ref{tab:word_specificity} shows the results of the term specificity experiments described in Section \ref{sec:word_specificity} measuring information content.
For both nouns and verbs, none of the diagnoses had significantly different term specificity levels compared to the control group in both the Twitter-STMHD and eRisk datasets. This verifies that the significant differences between the MHCs and the control group for emotion granularity is not due to varying word specificity levels in these groups.

\begin{table*}[t]
\centering
{\small
\begin{tabular}{lllrrr}
\hline
&  &  &
  \textbf{df} &
  \textbf{T-Statistic} &
  \textbf{P-value} \\ 
 \hline
 \textbf{POS} & \textbf{Dataset} & \textbf{MHC--Control} & & & \\ \hline
Noun & Twitter-STMHD & ADHD--control & 2368.64 & -1.94 & 0.144\\
& & Anxiety--control & 2233.36 & -0.58 & 0.718\\   
& & Bipolar--control & 1726.26 & -2.40 & 0.131\\   
& & MDD--control & 226.83 & -0.52 & 0.718\\ 
& & OCD--control & 1817.93 & 1.93 & 0.144\\ 
& & PPD--control & 178.33 & -0.49 & 0.718\\  
& & PTSD--control & 2237.82 & -0.54 & 0.718\\   
& & Depression--control & 2245.64 & -0.27 & 0.787\\  
& Reddit eRisk & Depression--control & 128.95 & 0.98 & 0.330 \\ \hline
Verb & Twitter-STMHD & ADHD--control & 2248.45 & -0.73 & 0.530\\
& & Anxiety--control & 2235.85 & 1.12 & 0.420\\   
& & Bipolar--control & 1852.44 & -2.10 & 0.096 \\   
& & MDD--control & 213.0 & 1.36 & 0.354\\ 
& & OCD--control & 1645.17 & 2.28 & 0.091\\ 
& & PPD--control & 169.49 & 0.98 & 0.438\\  
& & PTSD--control & 2351.12 & 2.53 & 0.091\\   
& & Depression--control & 2274.59 & 0.54 & 0.589\\  
& Reddit eRisk & Depression--control & 110.40 & 1.59 & 0.116\\
\hline 
\end{tabular}
}
\caption{ \textbf{Information Content}: 
The degrees of freedom, t-statistic and p-value for the word specificity experiments described in Section \ref{sec:word_specificity}. 
}
\label{tab:word_specificity}
\end{table*}

\section{Emotion Correlations}
Table \ref{tab:group_emo_corr} shows the group-averaged Spearman correlations for emotion pairs in the positive, negative, mixed valence groups, and the within-group and cross-group averages, for the Control groups, and the delta from these values for each MHC in both datasets.

Table \ref{tab:pairwise_emo_corr} shows the Spearman correlation between emotion arcs for all pairs of emotions for the control group. These results indicate that as baselines largely emotions in the same group (e.g., positive, negative, mixed, overall) co-occur more often than emotions across groups.

\begin{table*}
    \centering
    \small
    \begin{tabular}{lrrrrr}
\toprule
\textbf{Dataset, MHC} & $EG(pos)$ & $EG(neg)$ & $EG(var)$ & $EG(cross)$ & $EG(overall)$ \\
\midrule
\textit{Twitter-STMHD} \\
$\;\;\;$ Control & 0.027 & 0.023 & 0.012 & 0.006 & 0.022 \\
\hline
$\;\;\;$ ADHD & -0.012* & -0.008* & -0.005* & -0.010* & -0.010* \\
$\;\;\;$ Anxiety & -0.012* & -0.008* & -0.003* & -0.008* & -0.010* \\
$\;\;\;$ Bipolar & -0.004* & -0.008* & -0.004* & -0.002* & -0.006* \\
$\;\;\;$ MDD & -0.011* & -0.002 & -0.001 & -0.005* & -0.005* \\
$\;\;\;$ OCD & -0.013* & -0.009* & -0.006* & -0.009* & -0.009* \\
$\;\;\;$ PPD & -0.005 & -0.009* & -0.005 & -0.004 & -0.003 \\
$\;\;\;$ PTSD & -0.013* & -0.014* & -0.006* & -0.008* & -0.013* \\
$\;\;\;$ Depression & -0.011* & -0.005* & -0.003* & -0.005* & -0.008* \\
\midrule
\textit{Reddit eRisk} \\
$\;\;\;$Control & 0.114 & 0.117 & 0.090 & 0.094 & 0.112 \\
\hline
$\;\;\;$Depression & -0.016* & -0.021* & -0.012 & -0.022* & -0.017* \\
\bottomrule
\end{tabular}
\caption{\textbf{Emotion Granularity - Spearman correlations:}
 Spearman correlation values between utterance-level emotion arcs for the Control group, and the delta for each MHC when compared to the Control group. Emotion granularity is defined as the negative of these correlations (i.e, higher correlations imply a lower granularity). 
 Hyperparameters are the same as in Table \ref{tab:arrows_all_emotions_no_comorbid_zero-50-25_ex}.
}
\label{tab:group_emo_corr}
\end{table*}

\begin{table*}[t]
\centering
{\small
\begin{tabular}{lllllllll}
\toprule
 & anger & anticipation & disgust & fear & joy & sadness & surprise & trust \\
\midrule
anger & -- & -0.003 & 0.020 & 0.027 & -0.010 & 0.024 & 0.009 & 0.007 \\
anticipation & -- & -- & -0.003 & 0.007 & 0.021 & 0.004 & 0.012 & 0.027 \\
disgust & -- & -- & -- & 0.023 & -0.010 & 0.021 & 0.006 & 0.003 \\
fear & -- & -- & -- & -- & -0.003 & 0.021 & 0.012 & 0.013 \\
joy & -- & -- & -- & -- & -- & -0.000 & 0.008 & 0.027 \\
sadness & -- & -- & -- & -- & -- & -- & 0.012 & 0.008 \\
surprise & -- & -- & -- & -- & -- & -- & -- & 0.023 \\
\bottomrule
\end{tabular}
}
\caption{ \textbf{Emotion--Emotion Spearman correlations}: 
 Spearman correlation values between pairs of utterance-level emotion arcs for the all users in the control group of the Twitter-STMHD dataset. Hyperparameters are the same as in Table \ref{tab:arrows_all_emotions_no_comorbid_zero-50-25_ex}.
}
\label{tab:pairwise_emo_corr}
\end{table*}

\end{document}